\documentclass[runningheads]{llncs}

\usepackage{graphicx}
\usepackage{mathtools}
\usepackage{amsmath}
\usepackage{amssymb}
\usepackage{wrapfig}
\usepackage{booktabs}
\usepackage{multirow}
\usepackage{subcaption} 
\usepackage[table]{xcolor}

\begin{document}
\title{Learning Expected Appearances for Intraoperative Registration during Neurosurgery}
\titlerunning{Learning Expected Appearances for Intraoperative Registration}


\author{Nazim Haouchine\inst{1} \and
Reuben Dorent\inst{1} \and 
Parikshit Juvekar\inst{1}\and
Erickson Torio\inst{1}\and
William M. Wells III\inst{1,2}\and
Tina Kapur\inst{1}\and
Alexandra J. Golby\inst{1}\and
Sarah Frisken\inst{1}}


\authorrunning{Haouchine et al.}


\institute{Harvard Medical School, Brigham and Women's Hospital, Boston, MA, USA \and Massachusetts Institute of Technology, Cambridge, MA, USA
}

\maketitle              
\begin{abstract}
We present a novel method for intraoperative patient-to-image registration by learning Expected Appearances. 
Our method uses preoperative imaging to synthesize patient-specific expected views through a surgical microscope for a predicted range of transformations. 
Our method estimates the camera pose by minimizing the dissimilarity between the intraoperative 2D view through the optical microscope and the synthesized expected texture. 
In contrast to conventional methods, our approach transfers the processing tasks to the preoperative stage, reducing thereby the impact of low-resolution, distorted, and noisy intraoperative images, that often degrade the registration accuracy.
We applied our method in the context of neuronavigation during brain surgery. 
We evaluated our approach on synthetic data and on retrospective data from 6 clinical cases. 
Our method outperformed state-of-the-art methods and achieved accuracies that met current clinical standards.

\end{abstract}

\section{Introduction}
\label{sec:intro}
We address the important problem of intraoperative patient-to-image registration in a new way by relying on preoperative data to synthesize plausible transformations and appearances that are \textit{expected} to be found intraoperatively.
In particular, we tackle intraoperative 3D/2D registration during neurosurgery, where preoperative MRI scans need to be registered with intraoperative surgical views of the brain surface to guide neurosurgeons towards achieving a maximal safe tumor resection \cite{Sanai}.
Indeed, the extent of tumor removal is highly correlated with patients’ chances of survival and complete resection must be balanced against the risk of causing new neurological deficits \cite{Gonzalez-Darder2019} making accurate intraoperative registration a critical component of neuronavigation.

Most existing techniques perform patient-to-image registration using intraoperative MRI \cite{Kuhnt_2012}, CBCT \cite{Pereira} or ultrasound \cite{JI}\cite{Mohammadi}\cite{Rivaz}. For 3D-3D registration, 3D shape recovery of brain surfaces can be achieved using near-infrared cameras \cite{Filipe}, phase-shift 3D shape measurement \cite{Nakajima}, pattern projections \cite{Mohammadi} or stereovision \cite{JI20141169}.
The 3D shape can subsequently be registered with the preoperative MRI using conventional point-to-point methods such as iterative closest point (ICP) or coherent point drift (CPD). 
Most of these methods rely on cortical vessels that bring salient information for such tasks.
For instance, in \cite{Haouchine2020}, cortical vessels are first segmented using a deep neural network (DNN) and then used to constrain a 3D/2D non-rigid registration. 
The method uses physics-based modeling to resolve depth ambiguities. 
A manual rigid alignment is however required to initialize the optimization.
Alternatively, cortical vessels have been used in \cite{Luo} where sparse 3D points, manually traced along the vessels, are matched with vessels extracted from the preoperative scans. 
A model-based inverse minimization problem is solved by estimating the model's parameters from a set of pre-computed transformations. 
The idea of pre-computing data for registration was introduced by \cite{Sun}, who used an atlas of pre-computed 3D shapes of the brain surface for registration.
In \cite{Haouchine2022}, a DNN is trained on a set of pre-generated preoperative to intraoperative transformations. 
The registration uses cortical vessels, segmented using another neural network, to find the best transformation from the pre-generated set. 

The main limitation of existing intraoperative registration methods is that they rely heavily on processing intraoperative images to extract image features (eg., 3D surfaces, vessels centerlines, contours, or other landmarks) to drive registration, making them subject to noise and low-resolution images that can occur in the operating room \cite{Stoyanov2012}\cite{MAIERHEIN2022102306}.
Outside of neurosurgery, the concept of pre-generating data for optimizing DNNs for intraoperative registration has been investigated for CT to x-ray registration in radiotherapy where x-ray images can be efficiently simulated from CTs as digital radiographic reconstructions \cite{Lecompte} \cite{LiftReg}. 
In more general applications, case-centered training of DNNs is gaining in popularity and demonstrates remarkable results \cite{nerf}.

\begin{figure}[ht!]
     \begin{center}
        \includegraphics[width=1\linewidth]{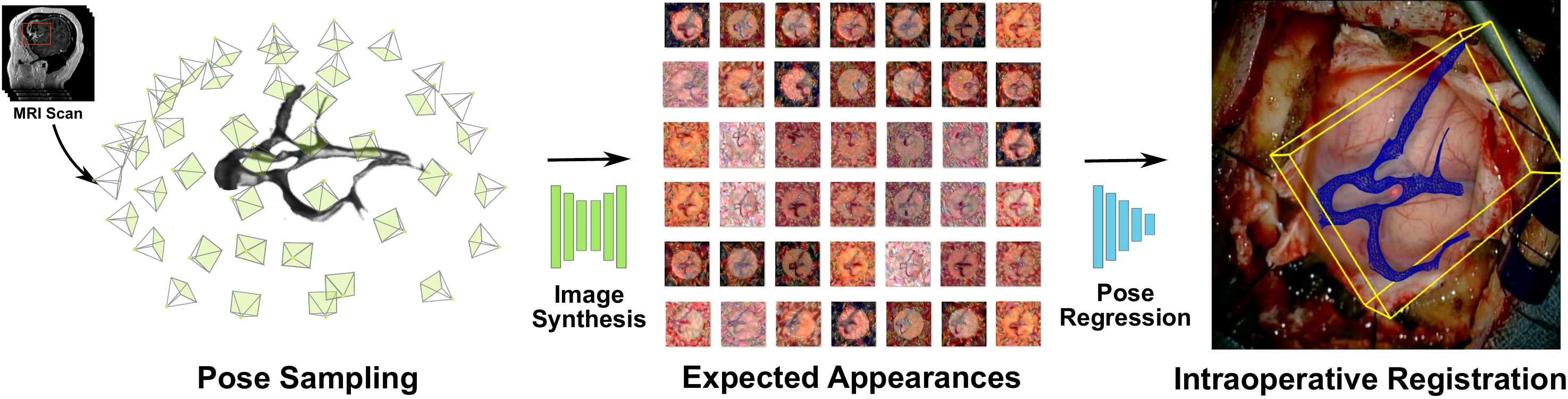}
    \end{center}
    \caption{Our approach estimates the 6-DoF camera pose that aligns a preoperative 3D mesh derived from MRI scans onto an intraoperative RGB image acquired from a surgical camera. We optimize a regressor network $\mathcal{P}_{\Omega}$ over a set of Expected Appearances that are generated by first sampling multiple poses and appearances from the 3D mesh  using neural image analogy through $\mathcal{S}_{\Theta}$.}
    \label{fig:overview}
\end{figure}

\paragraph{\textbf{Contribution:}} 
We propose a novel approach for patient-to-image registration that registers the intraoperative 2D view through the surgical microscope to preoperative MRI 3D images by learning \textit{Expected Appearances}.
As shown in Fig. \ref{fig:overview}, we formulate the problem as a camera pose estimation problem that finds the optimal 3D pose minimizing the dissimilarity between the intraoperative 2D image and its pre-generated Expected Appearance.
A set of Expected Appearances are synthesized from the preoperative scan and for a set of poses covering the range of plausible 6 Degrees-of-Freedom (DoF) transformations.  
This set is used to train a patient-specific pose regressor network to obtain a model that is texture-invariant and is cross-modality to bridge the MRI and RGB camera modalities. 
Similar to other methods, our approach follows a monocular single-shot registration, eliminating cumbersome and tedious calibration of stereo cameras, the laser range ﬁnder, or optical trackers.
In contrast to previous methods, our approach does not involve processing intraoperative images which have several advantages: it is less prone to intraoperative image acquisition noise; it does not require pose initialization; and is computationally fast thus supporting real-time use.
We present results on both synthetic and clinical data and show that our approach outperformed state-of-the-art methods.

\section{Method}
\subsection{Problem Formulation}
As illustrated in Fig. \ref{fig:overview}, given a 3D surface mesh of the cortical vessels $\mathbf{M}$, derived from a 3D preoperative scan, and a 2D monocular single-shot image of the brain surface $\mathbf{I}$, acquired intraoperatively by a surgical camera, we seek to estimate the 6-DoF transformation that aligns the mesh $\mathbf{M}$ to the image $\mathbf{I}$. 
Assuming a set of 3D points $\mathbf{u} = \{ u_j \in \mathbb{R}^3 \}\subset\mathbf{M}$ and a set of 2D points in the image $\mathbf{v} = \{ v_i \in \mathbb{R}^2 \}\subset\mathbf{I}$, solving for this registration problem can be formalized as finding the 6-DoF camera pose that minimizes the reprojection error: 
$ \sum_{i}^{n_c} \lVert \mathbf{A}[\mathbf{R}|\mathbf{t}] \begin{bsmallmatrix} u_{c_i}^w \\ 1 \end{bsmallmatrix} - v_i \lVert_2^2$, where $\mathbf{R} \in SO(3)$ and $\mathbf{t} \in \mathbb{R}^3$ represent a 3D rotation and 3D translation, respectively, and $\mathbf{A}$ is the camera intrinsic matrix composed of the focal length and the principal points (center of the image) while $\{c_i\}_i$ is a correspondence map and is built so that if a 2D point $v_i$ corresponds to a 3D point $u_j$ where $c_i = j$ for each point of the two sets.
Note that the set of 3D points $\mathbf{u}$ is expressed in homogenous coordinates in the minimization of the reprojection error.

In practice, finding the correspondences set $\{c_i\}_i$ between $\mathbf{u}$ and $\mathbf{v}$ is non-trivial, in particular when dealing with heterogeneous preoperative and intraoperative modality pairs (MRI, RGB Cameras, ultrasound, etc) which is often the case in surgical guidance. 
Existing methods often rely on feature descriptors \cite{Machado}, anatomical landmarks \cite{Luo}, or organ's contours and segmentation \cite{Haouchine2020}\cite{nercessian2021deep} involving tedious processing of the intraoperative image that is sensitive to the computational image noise.
We alleviate these issues by directly minimizing the dissimilarity between the image $\mathbf{I}$ and its Expected Appearance synthesized from $\mathbf{M}$.
By defining a synthesize function $\mathcal{S}_{\Theta}$ that synthesizes a new image $\widehat{\mathbf{I}}$ given a projection of a 3D surface mesh for different camera poses, i.e. $\widehat{\mathbf{I}} = \mathcal{S}_{\Theta}(\mathbf{A}[\mathbf{R}|\mathbf{t}], \mathbf{M})$, the optimization problem above can be rewritten as: 
\begin{equation}
\underset{\mathbf{A[R|t]}}{\mathrm{argmin}} \; \Big\{
\underset{\Theta}{\mathrm{min}}
\big\Vert \mathbf{I} - \mathcal{S}_{\Theta}\big(\mathbf{A}[\mathbf{R}|\mathbf{t}], \mathbf{M}\big) \big\Vert \Big \}
\label{eq:global_min}   
\end{equation}
This new formulation is correspondence-free, meaning that it alleviates the requirement of the explicit matching between $\mathbf{u}$ and $\mathbf{v}$. 
This is one of the major strengths of our approach. 
It avoids the processing of $\mathbf{I}$ at run-time, which is the main source of registration error. 
In addition, our method is patient-specific, centered around $\mathbf{M}$, since each model is trained specifically for a given patient. 
These two aspects allow us to transfer the computational cost from the intraoperative to the preoperative stage thereby optimizing intraoperative performance.   
The following describes how we build the function $\mathcal{S}_{\Theta}$ and how to solve Eq. \ref{eq:global_min}.

\begin{figure}[h!]
    \begin{center}
        \includegraphics[width=0.75\linewidth]{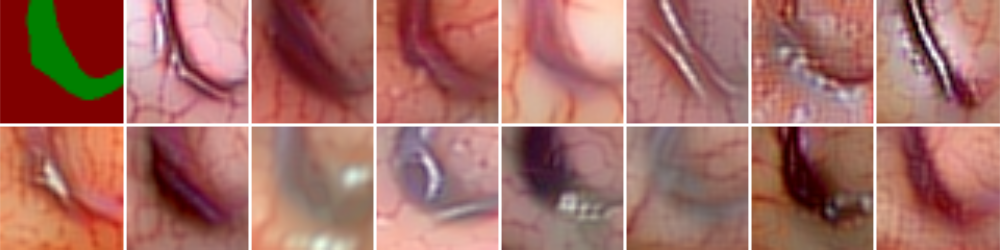}
    \end{center}
    \caption{An example of a set of Expected Appearances showing the cortical brain surface with the parenchyma and vessels. The network $\mathcal{S}_\Theta$ uses the binary image (top-left corner) computed from projecting $\mathbf{M}$ using $\mathbf{[R|t]}$ to semantically transfer 15 different textures $\{\mathbf{T}\}$ and synthesize the Expected Appearances $\{\widehat{\mathbf{I}}\}$.}
    \label{fig:nst}
\end{figure}

\subsection{Expected Appearances Synthesis}
We define a synthesis network $\mathcal{S}_\Theta : (\mathbf{A[R|t]}, \mathbf{M}, \mathbf{T}) \rightarrow \widehat{\mathbf{I}}$, that will generate a new image resembling a view of the brain surface from the 2D projection of the input mesh $\mathbf{M}$ following $\mathbf{[R|t]}$, and a texture $\mathbf{T}$. 
Several methods can be used to optimize $\Theta$. However, they require a large set of annotated data \cite{GANs} \cite{spadeBRAIN} or perform only on modalities with similar sensors \cite{Lecompte} \cite{LiftReg}. 
Generating RGB images from MRI scans is a challenging task because it requires bridging a significant difference in image modalities. 
We choose to use a neural image analogy method that combines the texture of a source image with a high-level content representation of a target image without the need for a large dataset \cite{ulyanov16}.
This approach transfers the texture from $\mathbf{T}$ to $\widehat{\mathbf{I}}$ constrained by the projection of $\mathbf{M}$ using $\mathbf{A[R|t]}$ by minimizing the following loss function:
\begin{equation}
\mathcal{L}_\Theta = \sum_{l} \sum_{ij} 
\Big(
\mathbf{w}^{l,c}_{\mathbf{T}_\text{class}} \mathcal{G}_{ij}^l(\mathbf{T}) - \mathbf{w}^{l,c}_{\mathbf{A[R|t],M}} \mathcal{G}_{ij}^l(\widehat{\mathbf{I}}) 
\Big) \quad \text{for} \; c \in \{0,1,2\}   
\end{equation}
where $\mathcal{G}_{ij}^l(\mathcal{T})$ is the Gram matrix of texture $\mathcal{T}$ at the $l$-th convolutional layer (pre-trained VGG-19 model), and $\mathbf{w}^{l,c}_{\mathbf{T}_\text{class}}$ are the normalization factors for each Gram matrix, normalized by the number of pixels in a label class $c$ of $\mathbf{T}_\text{class}$. 
This allows for the quantification of the differences between the texture image $\mathbf{T}$ and the generated  image $\widehat{\mathbf{I}}$ as it is being generated. 
Importantly, computing the inner-most sum over each label class $c$ allows for texture comparison within each class, for instance: the background, the parenchyma, and the cortical vessels.

In practice, we assume constant camera parameters $\mathbf{A}$ and first sample a set of binary images by randomly varying the location and orientation of a virtual camera $\mathbf{[R|t]}$ w.r.t. to the 3D mesh $\mathbf{M}$ before populating the binary images with the textures using $\mathcal{S}_\Theta$ (see Fig. \ref{fig:overview}). 
We restrict this sampling to the upper hemisphere of the 3D mesh to remain consistent with the plausible camera positions w.r.t. patient's head during neurosurgery.

We use the L-BFGS optimizer and 5 convolutional layers of VGG-19 to generate each image following \cite{ulyanov16} to find the resulting parameters $\widehat{\Theta}$.
The training to synthesize for a single image typically takes around 50 iterations to converge.

\subsection{Pose Regression Network}
In order to solve Eq. \ref{eq:global_min}, we assume a known focal length that can be obtained through pre-calibration.
To obtain a compact representation of the rotation and since poses are restricted to the upper hemisphere of the 3D mesh (No Gimbal lock),  the Euler-Rodrigues representation is used.
Therefore, there are six parameters to be estimated: rotations $r_x, r_y, r_z$ and translations $t_x, t_y, t_z$.
We estimate our 6-DoF pose with a regression network $P_{\Omega} : \mathbf{I} \rightarrow \mathbf{p}$ and optimize its weights $\Omega$ to map each synthetic image $\mathbf{I}$ to its corresponding camera pose $\mathbf{p} = [r_x, r_y, r_z, t_x, t_y, t_z]^\mathsf{T}$.

The network architecture of $\mathcal{P}_\Omega$ consists of 3 blocks each composed of two convolutional layers and one $\textsf{ReLU}$ activation.
To decrease the spatial dimension, an average pooling layer with a stride of 2 follows each block except the last one.
At the end of the last hierarchy, we add three fully-connected layers with 128, 64, and 32 neurons and $\textsf{ReLU}$ activation followed by one fully-connected with 6 neurons with a linear activation.
We use the set of generated Expected Appearances $T^\mathcal{P} = \{(\mathbf{I}_i;\mathbf{p}_i)\}_i$; and optimize the following loss function over the parameters $\Omega$ of the network $\mathcal{P}_\Omega$ :
\begin{equation}
\mathcal{L}_\Omega = \sum_{i=1}^{|T^\mathcal{P}|} \Big (\big \Vert \mathbf{t}_i  - \widehat{\mathbf{t}}_i \big \Vert_2 + \Vert \mathbf{R}^{\text{vec}}_i  - \widehat{\mathbf{R}}^{\text{vec}}_i \big \Vert_2 \Big )
\end{equation}
where $\mathbf{t}$ and $\mathbf{R}^{\text{vec}}$ are the translation and rotation vector, respectively. 
We experimentally noticed that optimizing these entities separately leads to better results. 
The model is trained for each case (patient) for 200 epochs using mini-batches of size 8 with Adam optimizer and a learning rate of $0.001$ and decays exponentially to $0.0001$ over the course of the optimization.
Finally, at run-time, given an image $\mathbf{I}$ we directly predict the corresponding 3D pose $\mathbf{p}$ so that: $\mathbf{p} \leftarrow \mathcal{P}(\mathbf{I}; \widehat{\Omega})$, where $\widehat{\Omega}$ is the resulting parameters from the training.

\section{Results}
\begin{figure}[t!]
\centering 
 \subfloat{\includegraphics[width=0.164\linewidth]{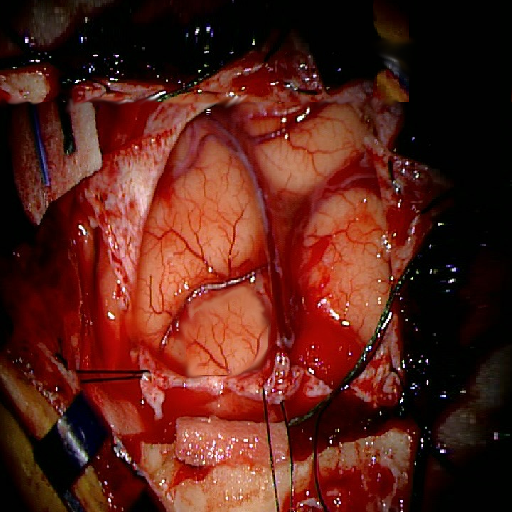}}
 \hfill
 \subfloat{\includegraphics[width=0.164\linewidth]{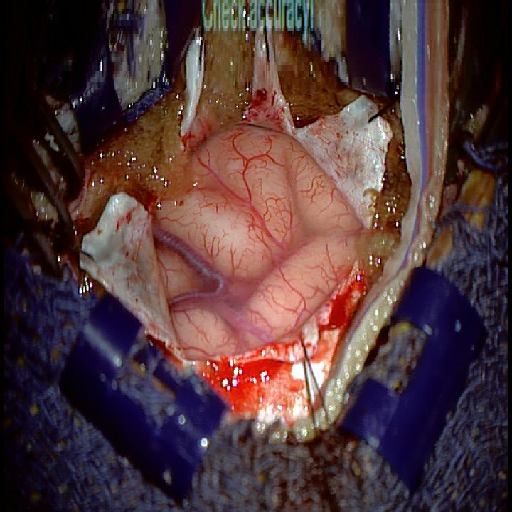}}
 \hfill
 \subfloat{\includegraphics[width=0.164\linewidth]{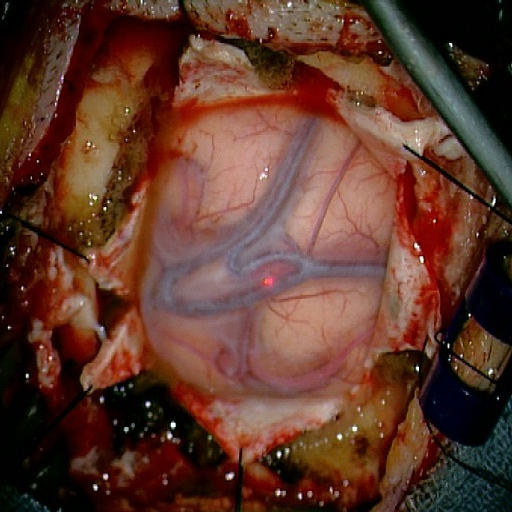}}
 \hfill
 \subfloat{\includegraphics[width=0.164\linewidth]{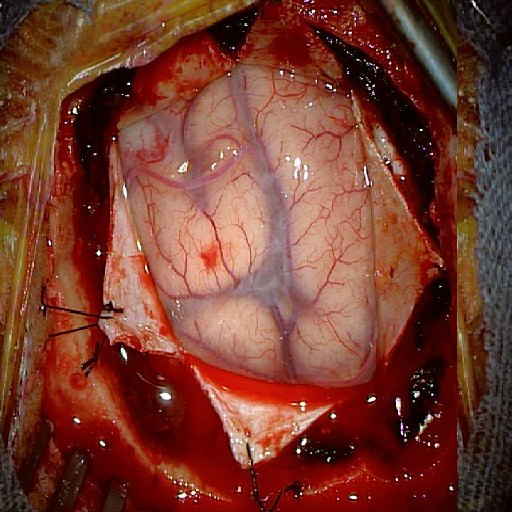}}
 \hfill
 \subfloat{\includegraphics[width=0.164\linewidth]{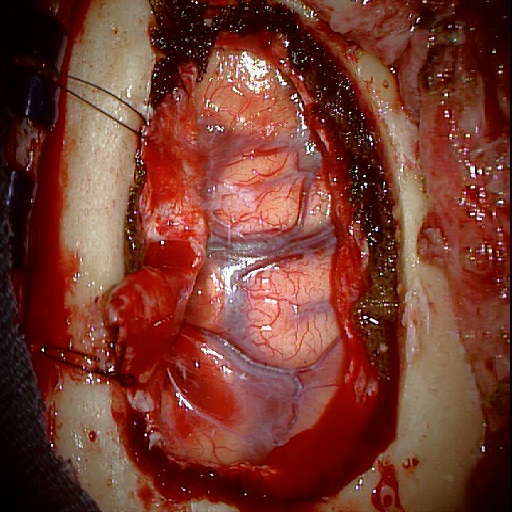}}
 \hfill
 \subfloat{\includegraphics[width=0.164\linewidth]{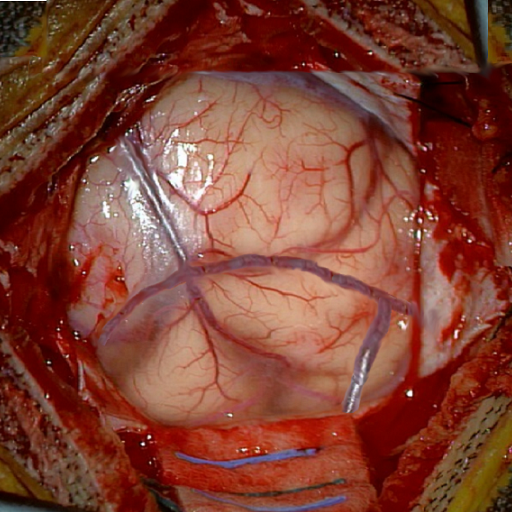}} \\

 \subfloat{\includegraphics[width=0.164\linewidth]{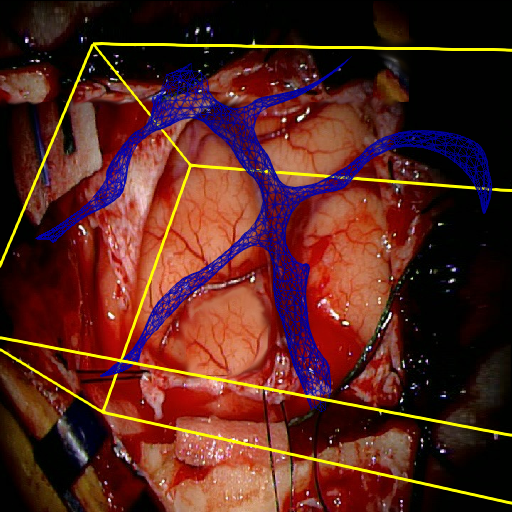}}
 \hfill
 \subfloat{\includegraphics[width=0.164\linewidth]{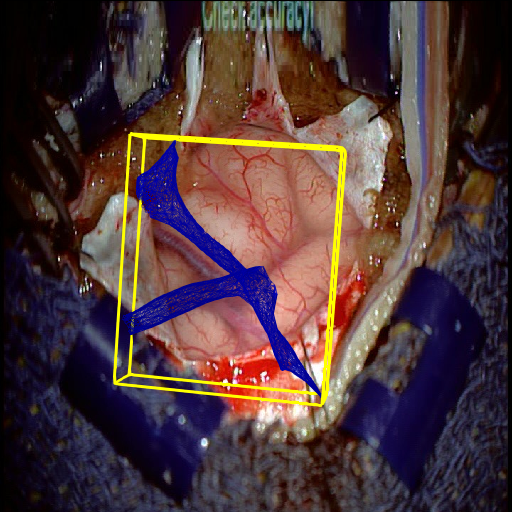}}
 \hfill
 \subfloat{\includegraphics[width=0.164\linewidth]{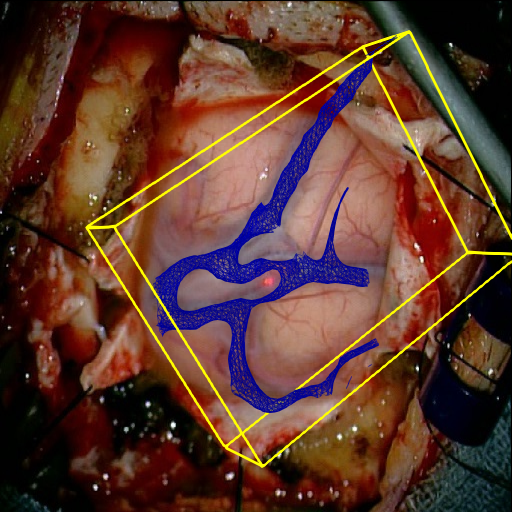}}
 \hfill
 \subfloat{\includegraphics[width=0.164\linewidth]{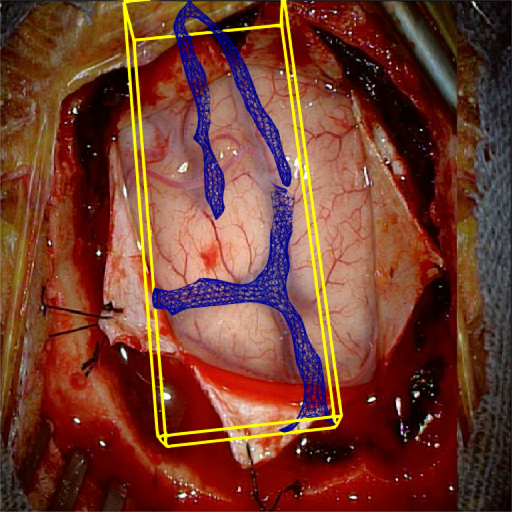}}
 \hfill
 \subfloat{\includegraphics[width=0.164\linewidth]{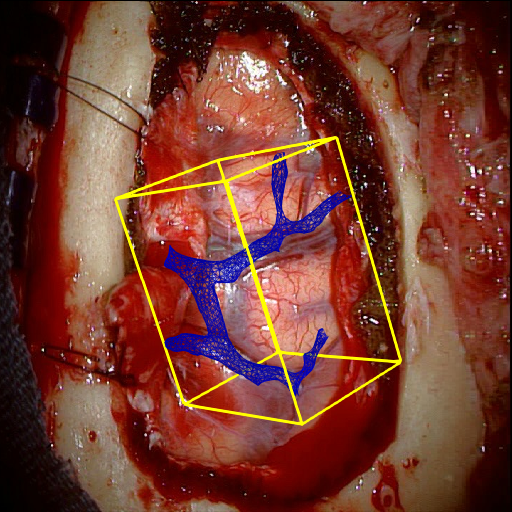}}
 \hfill
 \subfloat{\includegraphics[width=0.164\linewidth]{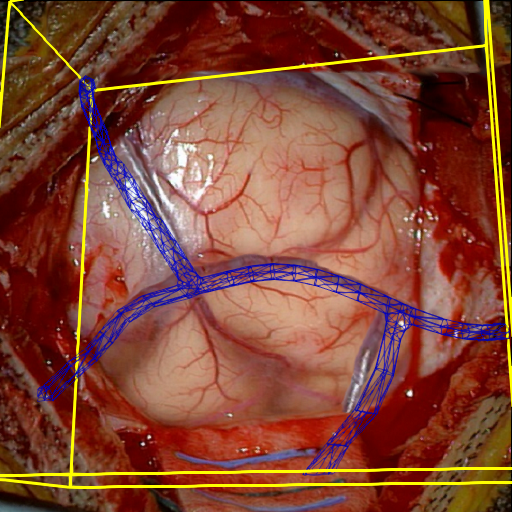}} \\
 
 \subfloat{\includegraphics[width=0.164\linewidth]{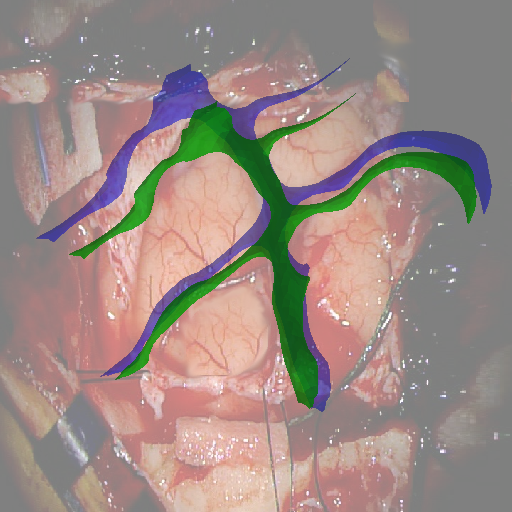}}
 \hfill
 \subfloat{\includegraphics[width=0.164\linewidth]{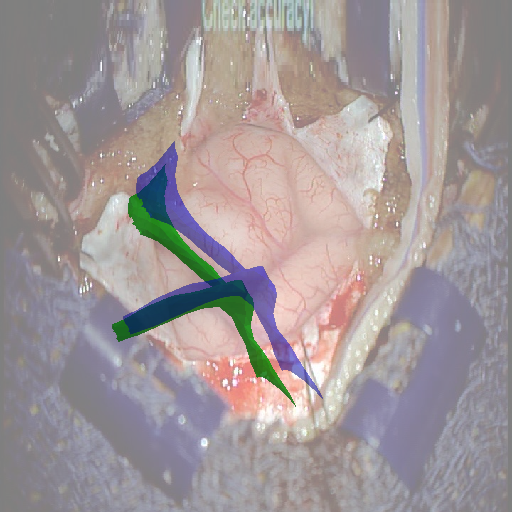}}
 \hfill
 \subfloat{\includegraphics[width=0.164\linewidth]{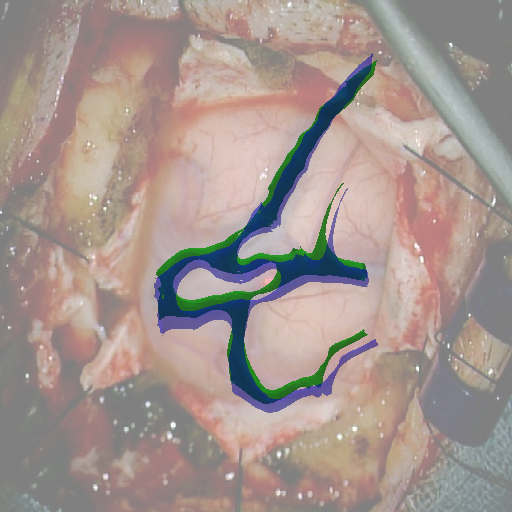}}
 \hfill
 \subfloat{\includegraphics[width=0.164\linewidth]{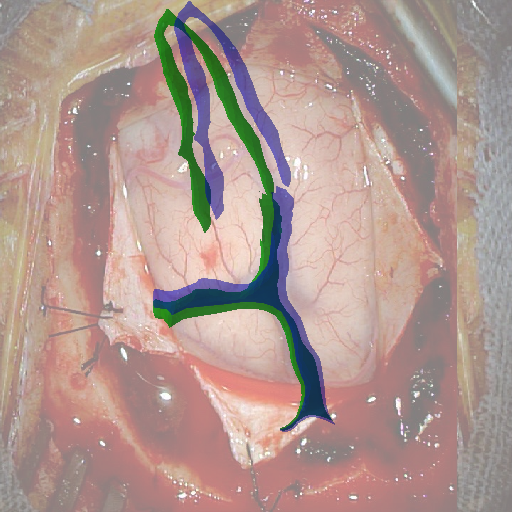}}
 \hfill
 \subfloat{\includegraphics[width=0.164\linewidth]{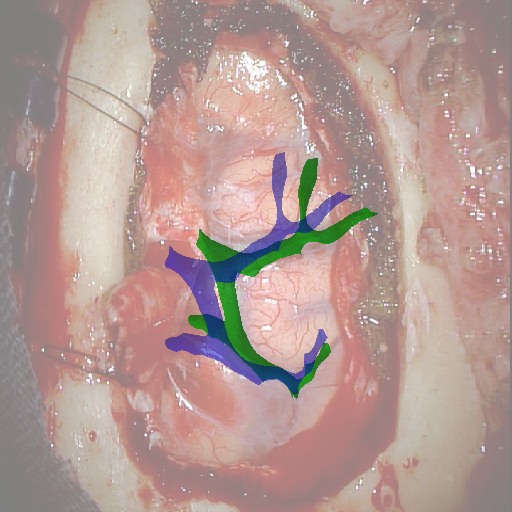}}
 \hfill
 \subfloat{\includegraphics[width=0.16\linewidth]{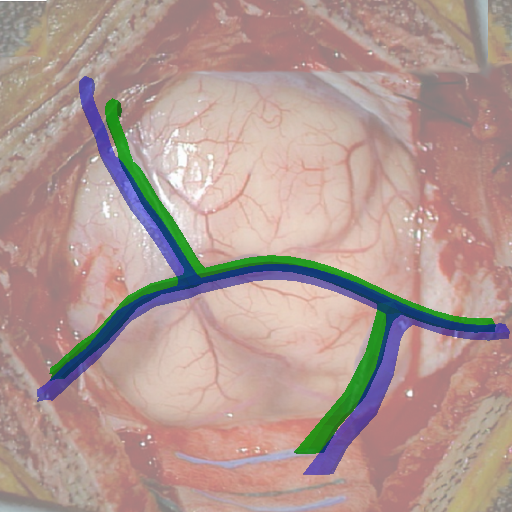}}
 \caption{Qualitative results on 6 patient datasets retrospectively showed in the first row. The second row shows Augmented Reality views with our predicted poses. The third row highlights 3D mesh-to-image projections using the ground-truth poses (green), and our predicted poses (blue). Our predictions are close to the ground truth for all cases. Note: microscope-magnified images with visible brain surface diameter $\approx$35\,mm.}
\label{fig:ar}
\vspace*{-1em}
\end{figure}

\vspace{-1pt}
\paragraph{\textbf{Dataset}} We tested our method retrospectively on 6 clinical datasets from 6 patients (cases) (see Fig. \ref{fig:ar}).
These consisted of preoperative T1 contrast MRI scans and intraoperative images of the brain surface after dura opening. Cortical vessels around the tumors were segmented and triangulated to generate 3D meshes using \textit{3D Slicer}.
We generated 100 poses for each 3D mesh (i.e.: each case) and used a total of 15 unique textures from human brain surfaces (different from our 6 clinical datasets) for synthesis using $\mathcal{S}_\Theta$. 
In order to account for potential intraoperative brain deformations \cite{Frisken} we augment the textured projection with elastic deformation \cite{unet} resulting in approximately 1500 images per case. 
The surgical images of the brain (left image of the stereoscopic camera) were acquired with a Carl Zeiss surgical microscope.
The ground-truth poses were obtained by manually aligning the 3D meshes on their corresponding images.

We evaluated the pose regressor network on both synthetic and real data.
The model training and validation were performed on the synthesized images while the model testing was performed on the real images.
Because a conventional train/validation/test split would lead to texture contamination, we created our validation dataset so that at least one texture is excluded from the training set. 
On the other hand, the test set consisted of the real images of the brain surface acquired using the surgical camera and are never used in the training.

\begin{wrapfigure}{r}{0.35\textwidth}
\vspace{-2em}
\raisebox{0pt}[\dimexpr\height-0.6\baselineskip\relax]{\includegraphics[width=0.34\textwidth]{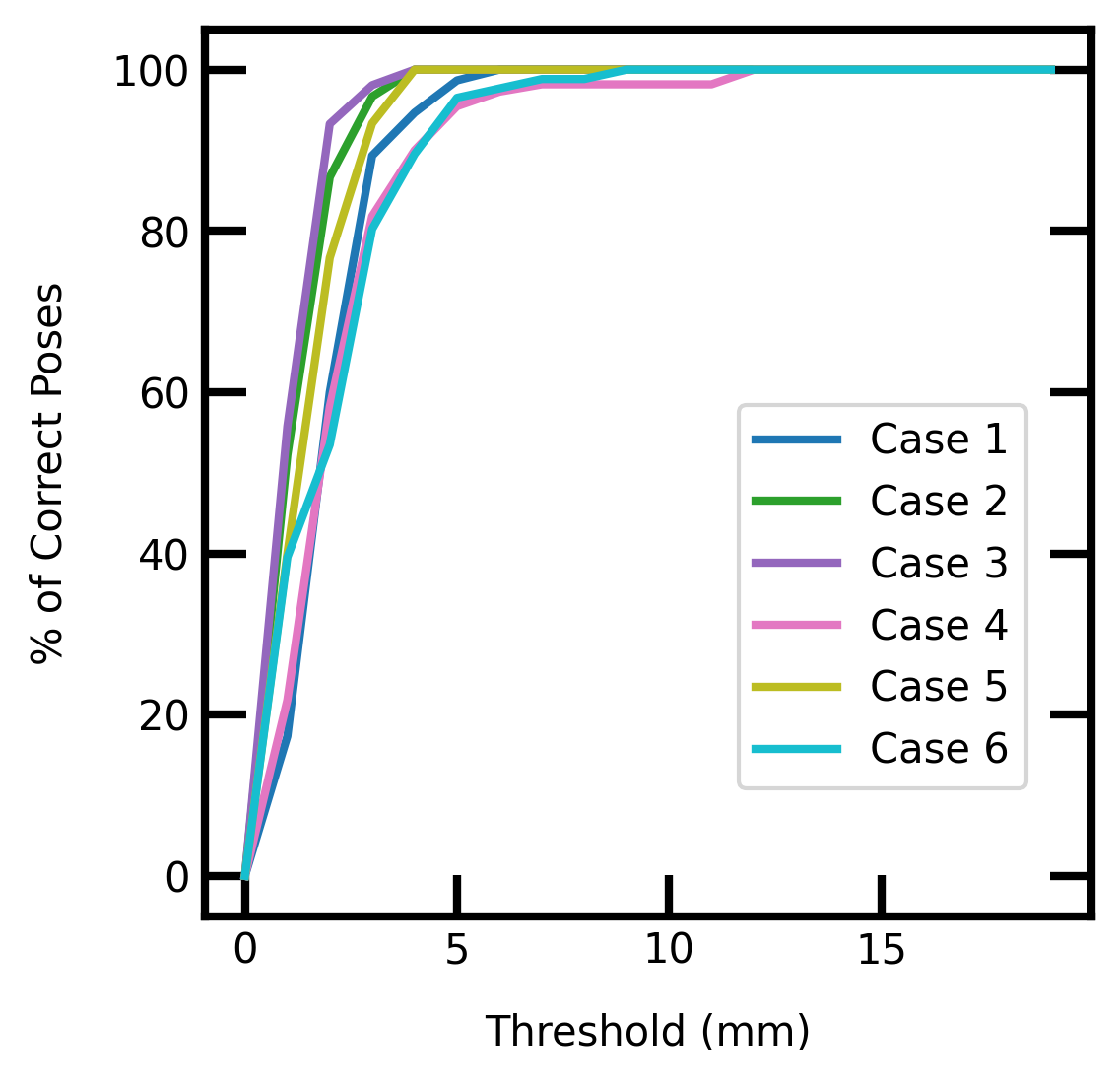}}
\caption{Accuracy-threshold curves on the validation set.}
\label{fig:atc}
\vspace{-2em}
\end{wrapfigure}


\paragraph{\textbf{Metrics}} We chose the average distance metric (ADD) as proposed in \cite{Shotton2013SceneCR} 
for evaluation. 
Given a set of mesh's 3D vertices, the ADD computes the mean of the pairwise distance between the 3D model points transformed using the ground truth and estimated transformation.
We also adjusted the default \textit{5cm-5deg} translation and rotation error to our neurosurgical application and set the new threshold to \textit{3mm-3deg}. 


\paragraph{\textbf{Accuracy-threshold Curves}} 
We calculated the number of 'correct' poses estimated by our model. 
We varied the distance threshold on the validation sets (excluding 2 textures) in order to reveal how the model performs w.r.t. that threshold.
We plotted accuracy-threshold curves showing the percentage of pose accuracy variation with a threshold in a range of $0$ mm to $20$ mm.
We can see in Fig. \ref{fig:atc} that a $80.23$\% pose accuracy was reached within the \textit{3mm-3deg} threshold for all cases. 
This accuracy increases to $95.45$\% with a \textit{5mm-5deg} threshold.

\begin{figure}[ht!]
\centering 
\includegraphics[width=1\linewidth]{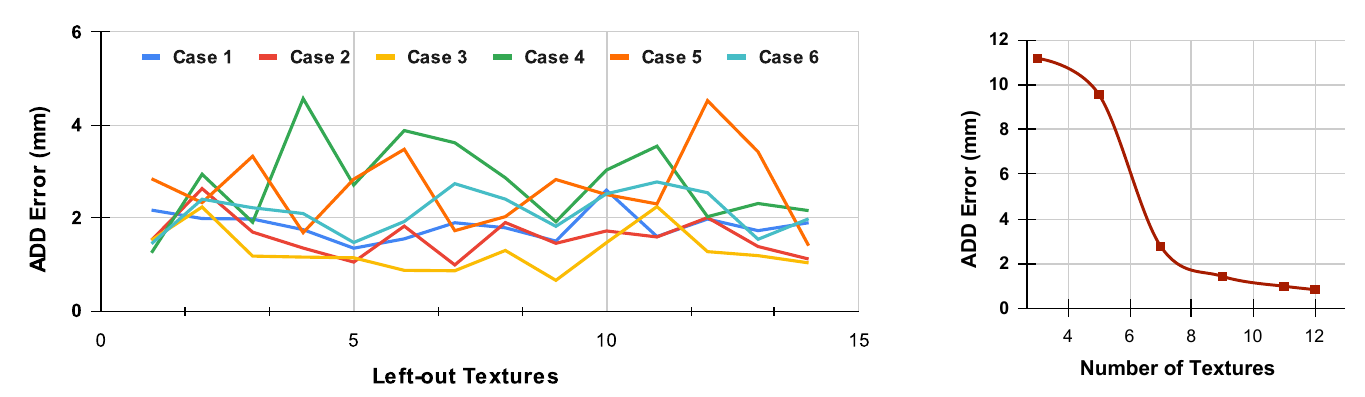}
 \caption{Evaluation of texture invariance: (left) Leave-one-\textit{Texture}-out cross validation and (right) impact of the number of textures on model accuracy.}
\label{fig:charts}
\end{figure}

\paragraph{\textbf{Validation and Evaluation of Texture Invariance}}
We chose to follow a Leave-one-\textit{Texture}-out cross-validation strategy to validate our model.
This strategy seemed the most adequate to prevent over-fitting on the textures.
We measured the ADD errors of our model for each case and report the results in Fig. \ref{fig:charts}-(left) for each left-out texture. 
We observed a variance in the ADD error that depends on which texture is left out. 
This supports the need for varying textures to improve the pose estimation.  
However, the errors remain low, with a $2.01\pm0.58$ mm average ADD error, over all cases. 
The average ADD error per case (over all left-out textures) is reported in Tab. \ref{tab:k-folds}.
We measured the impact of the number of textures on the pose accuracy by progressively adding new textures to the training set, starting from 3 to 12 textures, while leaving 3 textures out for validation.
We kept the size of the training set constant to not introduce size biases.
Fig. \ref{fig:charts}-(right) shows that increasing the number and variation of textures improved model performances.


\def\arraystretch{0.8}
\rowcolors{2}{}{gray!10}
\begin{table}[h!]
\centering
\caption{Validation on synthetic data and comparisons using real data.}
\begin{tabular}{l | l | l | c c c c c c } 
 \toprule
 Data & Experiment & Metric & Case 1 & Case 2 & Case 3 & Case 4 & Case 5 & Case 6 \\
 \midrule
  \multirow{2}{*}{\textit{Synth.} \space} & LOTO CV & Avg. ADD (mm) \space  & 1.80 & 1.56 & 1.27 &	2.83 &	2.58 &	2.07 \\
 & Acc./Thresh. \space & 3mm-3deg (\%) & 89.33 & 96.66 & 98.07 & 81.81 & 93.33 & 80.23 \\ 
 \midrule
  
 \multirow{3}{*}{\textit{Real}} & Ours & ADD (mm)& \bf 3.64 & \bf 2.98 & \bf 1.62 & 4.83 & \bf 3.02 & \bf 3.32 \\
 & ProbSEG  & ADD (mm) &  4.49 & 3.82 & 3.12 & \bf 4.69 & 4.99 & 3.67 \\
 & BinSEG & ADD (mm) & 9.2 &	4.87 & 12.12 & 8.09 & 11.43 &	6.29 \\
 \bottomrule
\end{tabular}
\label{tab:k-folds}
\end{table}

\paragraph{\textbf{Test and Comparison on Clinical Images}}
We compared our method (Ours) with segmentation-based methods (ProbSEG) and (BinSEG) \cite{Haouchine2022}.
These methods use learning-based models to extract binary images and probability maps of cortical vessels to drive the registration.
We report in Tab. \ref{tab:k-folds} the distances between the ground truth and estimated poses. 
Our method outperformed ProbSEG and BinSEG with an average ADD error of $3.26\pm1.04$\,mm compared to $4.13\pm0.70$\,mm and $8.67\pm2.84$\,mm, respectively.
Our errors remain below clinically measured neuronavigation errors reported in \cite{Frisken}, in which a $5.26\pm0.75$ mm average initial registration error was measured in 15 craniotomy cases using intraoperative ultrasound.
Our method outperformed ProbSEG in 5 cases out of 6 and BinSEG in all cases and remained within the clinically measured errors \textit{without} the need to segment cortical vessels or select landmarks from the intraoperative image.  
Our method also showed fast intraoperative computation times. It required an average of only $45$ milliseconds to predict the pose (tested on research code on a laptop with NVidia GeForce GTX 1070 8GB without any speciﬁc optimization), suggesting a potential use for real-time temporal tracking. 

Fig. \ref{fig:ar} shows our results as Augmented Reality views with bounding boxes and overlaid meshes.
Our method produced visually consistent alignments for all 6 clinical cases without the need for initial registration.
Because our current method does not account for brain-shift deformation, our  method produced some misalignment errors.
However, in all cases, our predictions are similar to the ground truth.

\section{Discussion and Conclusion}
\paragraph{\textbf{Clinical Feasibility.}} We have shown that our method is clinically viable. 
Our experiments using clinical data showed that our method provides accurate registration without manual intervention, that it is computationally efficient, and it is invariant to the visual appearance of the cortex. 
Our method does not require intraoperative 3D imaging such as intraoperative MRI or ultrasound, which require expensive equipment and are disruptive during surgery. 
Training patient-specific models from preoperative imaging transfers computational tasks to the preoperative stage so that patient-to-image registration can be performed in near real-time from live video acquired from a surgical microscope.

\paragraph{\textbf{Limitations.}} The method presented in this paper is limited to 6-DoF pose estimation and does not account for deformation of the brain due to changes in head position, fluid loss, or tumor resection and assumes a known focal length. 
In the future, we will expand our method to model non-rigid deformations of the 3D mesh and to accommodate expected changes in zoom and focal depth during surgery. 
We will also explore how texture variability can be controlled and adapted to the observed image to improve model accuracy.

\paragraph{\textbf{Conclusion.}} We introduced Expected Appearances, a novel learning-based method for intraoperative patient-to-image registration that uses synthesized expected images of the operative field to register preoperative scans with intraoperative views through the surgical microscope. 
We demonstrated state-of-the-art, real-time performance on challenging neurosurgical images using our method. 
Our method could be used to improve accuracy in neuronavigation and in image-guided surgery in general.

\section*{Acknowledgement}
The authors were partially supported by the following National Institutes of Health grants: R01EB027134, R03EB032050, R01EB032387, and R01EB034223.


\bibliographystyle{splncs04}
\bibliography{mybib}

\end{document}